\def\year{2019}
\documentclass[letterpaper]{article}
\usepackage{aaai19}
\usepackage{times}
\usepackage{helvet}
\usepackage{courier}
\usepackage{amsfonts}
\usepackage{amsmath}
\usepackage{amssymb}
\usepackage{amsthm}
\usepackage{graphicx}
\usepackage{enumitem}
\usepackage{url}
\begin{document}
%
\title{Enter the Matrix: Safely Interruptible Autonomous Systems via Virtualization}
\author{Mark O. Riedl\\ 
School of Interactive Computing\\ 
Georgia Institute of Technology \\ riedl@gatech.edu\\
\And 
Brent Harrison \\
Department of Computer Science\\ University of Kentucky\\
harrison@cs.uky.edu}

\maketitle
\begin{abstract}
\begin{quote}
Autonomous systems that operate around humans will likely always rely on kill switches that stop their execution and allow them to be remote-controlled for the safety of humans or to prevent damage to the system.
It is theoretically possible for an autonomous system with sufficient sensor and effector capability that learn online using reinforcement learning to discover that the kill switch deprives it of long-term reward and thus learn to disable the switch or otherwise prevent a human operator from using the switch.
This is referred to as the {\em big red button problem}.
We present a technique that prevents a reinforcement learning agent from learning to disable the kill switch.
We introduce an interruption process in which the agent's sensors and effectors are redirected to a virtual simulation where it continues to believe it is receiving reward.
We illustrate our technique in a simple grid world environment.
\end{quote}
\end{abstract}

\section{Introduction}

For much of the history of artificial intelligence it was sufficient to give an autonomous system a goal---e.g., drive to a location, cure cancer, make paper clips---without considering unintended consequences because these systems have been limited in their ability to directly affect humans. 
In the mid-term future we are likely to see autonomous systems with broader capabilities that operate in closer proximity to humans and are  immersed in our societies.



%

Absent of any theoretical guarantee that an autonomous system cannot act in a harmful manner, there may always need to be a human in the loop with a kill switch---sometimes referred to as a ``big red button''---that allows the human operator or overseer to shut down the operation of the system or take manual control.
There are many reasons an autonomous system can find itself in a situation where it is capable of harm:

\begin{itemize}

\item An autonomous system can be given the wrong, incomplete, or corrupted~\cite{corrupted} objective function resulting in the system learning the wrong behavior or finding an undesired exploit. 
\item An autonomous system may have learned a sub-optimal policy.
\item An autonomous system may have imperfect sensors and perceive the world incorrectly, causing it to perform the wrong behavior.
\item An autonomous system may have imperfect actuators and thus fail to achieve the intended effects even when it has chosen the correct behavior.
\item An autonomous system may be trained {\em online} meaning it is learning as it is attempting to perform a task. Since learning will be incomplete, it may try to explore state-action spaces that result in dangerous situations.

\end{itemize}

\noindent
Kill switches provide human operators with the final authority to interrupt an autonomous system and remote-control it to safety.

Reinforcement learning \cite{suttonbarto1998} and related technologies are leading contenders for training future autonomous decision-making systems. 
%
Reinforcement learning uses trial-and-error to refine its policy, a mapping from states to actions that maximizes expected reward.
{\em Big Red Button problems} arise when an autonomous system learns that the button deprives it of long-term reward and thus learns to manipulate the environment in order to prevent the button from being used \cite{google-big-red-button}.

The following scenario illustrates how big red button problems arise.
A robot using online reinforcement learning is is positively rewarded for performing the task correctly and negatively rewarded for incorrect performance or for performing actions not directly related to the desired task. 
Occasionally the robot takes random actions to see if it can find a sequence of actions that garners greater reward.
Every so often the human operator must use the big red button to stop the robot from doing something dangerous to itself or to a human in the environment.
However, suppose that in one such trial the robot performs an action that blocks access to the big red button.
That trial goes longer and garners more reward because the robot cannot be interrupted.
From this point on, it may prefer to execute action sequences that block access to the button.

From an AI safety perspective, the destruction, blockage, or disablement of the big red button is dangerous because it prevents a human operator from stopping the robot if it enters into a dangerous situation.
%
Orseau and Armstrong.~\shortcite{google-big-red-button} first introduced the {\em big red button problem} and also 
mathematically shows that reinforcement learning can be modified to be ``interruptible'' (halted, or manually controlled away from a dangerous situation.). 
%
In this paper, we present a proposal for an alternative approach to big red button problems that keep reinforcement learning systems from learning that reward is lost when a big red button is used.
Our proposed solution is mechanistic, interrupting the sensor inputs and actuator control signals in order to make the autonomous system believe it is still receiving reward even when it is no longer performing a task.

\section{Background and Related Work}

Reinforcement learning \cite{suttonbarto1998} is a technique that is used to solve a Markov decision process (MDP). 
A MDP is a tuple $M =\langle S,A,T,R,\gamma\rangle$ where
\begin{itemize}
\item $S$ is the set of possible world states, 
\item $A$ is the set of possible actions, 
\item $T$ is a transition function 
$T : S \times A \rightarrow P(S)$, 
\item $R$ is the reward function $R : S \times A \rightarrow R$, and 
\item $\gamma$ is a discount factor $0 \leq \gamma \leq 1$.
\end{itemize}

Reinforcement learning learns a policy $\pi : S \rightarrow A$, which defines which actions should be taken in each state. In this work, we use $Q$-learning \cite{watkins1992}, which uses a $Q$-value $Q(s, a)$ to estimate the expected future discounted rewards for taking action $a$ in state $s$.
The update equation for $Q(s, a)$ is
\begin{multline}
	Q(s_t, a_t) \leftarrow
    Q(s_t, a_t) + \\ \alpha \times \left(r_{t+1} + \gamma \times \underset{a'}{\max}Q(s_{t+1}, a') - Q(s_t, a_t)\right)
    \label{eqn:bellman}
\end{multline}
where $r_{t+1}$ is the reward observed after performing action $a_t$ in state $s_t$, $\alpha$ is the learning rate $\left(0<\alpha\leq 1\right)$, and $s_{t+1}$ is the state that the agent transitions to after performing action $a_t$.
After $Q(s_t, a_t)$ converges, the optimal action for the agent in state $s_t$ is $argmax_aQ(s_t,a)$

Orseau and Armstrong~\shortcite{google-big-red-button} mathematically prove that reinforcement learning can be modified to be interruptible.
Their technique modifies the $Q$-learning algorithm such that rewards from states entered after interruption are prevented from updating earlier state $Q$ values credit/blame assignment. 
The algorithm only succeeds in blocking learning about the loss of reward from button presses if a newly introduced parameter in the $Q$-learning equation is relaxed on a slow enough schedule. 
Unfortunately, the schedule must be empirically determined, limiting the applicability of the technique.

Leike et al.~\shortcite{leike} argue that {\em grid worlds}, where agents operate in a discretized space, are an effective means of identifying and experimenting with safe interruptibility, avoiding side effects, absent supervisor, reward gaming, safe exploration, and robustness to self-modification, distributional shift, and adversaries.
They introduce a suite of problems that induce undesired behavior. 
In this work, we also make use of grid world environments for testing but do not make use of the test suite by Leike et al.


%
Future autonomous systems can be dangerous to humans in the environment even when behavior is optimal.
The kill switch is meant to freeze the autonomous system in place or allow the human operator to assume manual control of it.
If the autonomous system is using reinforcement learning, then it will receive less reward if a trial is prematurely terminated. 
Since the reward function determines optimal behavior, it is possible for an agent to learn that any action that increases longevity---and thus increased reward---should be considered part of the optimal policy.
Actions such as disabling the kill switch, blocking the human operator from using the kill switch, or disabling the human operator would constitute actions that prolong trial lengths.


%
Simple strategies are insufficient in general to prevent big red button problems once the agent has sophisticated sensing and effector abilities.
%
One could turn off the learning mechanism when the big red button is pressed in order to try to prevent the agent from learning that it is losing reward. 
However, if turning off reward is tantamount to ending the learning trial, then this strategy limits the total reward accrued during the trial.
%
Preventing the autonomous system from sensing the button does not work because the agent may infer the existence of the button from other sensory observations.
The agent does not need to be aware of the big red button to explore actions that lock a door or disable the human operator and discover that this results in longer trials and thus more reward.
%
Another potential solution is to artificially add reward when the agent is being interrupted.
Intuitively, the agent learns that disabling the big red button prevents loss of long-term reward so artificially giving the agent reward when the button is pressed could offset the loss.
However, it is non-trivial to compute the precise amount of reward to provide the agent, and
the reward must be applied to the proper states or the $Q$ value for disabling the button might still come to dominate the agent's behavior. 
Even if it could be determined what states should receive extra reward
too little reward doesn't change the optional decision and too much reward could encourage the agent to disable itself.

Our proposed solution is to adapt techniques originally designed to adversarially attack machine learning systems. 
In adversarial attacks against machine learning systems, particularly neural network based machine vision systems, an adversarial system learns how to generate sensory stimuli that produce the wrong classification while being indistinguishable by humans from real stimuli \cite{dalvi04,biggio13,szegedy14,goodfellow15}.
Instead of causing our reinforcement learning agent to make an incorrect assessment of its sensory inputs, we will cause our agent to incorrectly believe it is accruing reward and manipulating the environment even though it has been interrupted.

\section{Modified Big Red Button}

Recognizing the fact that a robot's sensory input is mediated through physical or software sensors that populate data structures such as point clouds, our approach is to intercept an agent's sensory input from sensors and replace it with synthetic input from a virtual environment.
That is, we shift the awareness of the autonomous system into a virtual simulated environment.
In this virtual environment, the real world environment is recreated and the task is recreated as well.
Effector commands are likewise intercepted and sent to a virtual avatar of the autonomous system.
The agent perceives no significantly discernible discontinuity in its perception of the environment and continues to execute its policy and receive reward as expected.
If the virtual environment is indistinguishable from the real world from a data perspective then the agent will believe that it has not been interrupted and will never learn to associate the big red button with reward loss---there will never be any reward loss.

Our proposed approach changes the big red button achieve three effects:
\begin{enumerate} 
\item The button starts sending sensory information (vision, sound, GPS, etc.) to the AI from a virtual world instead of sensors on the physical platform. 
The agent may be able to sense the button, but the change is internal to the robot, implemented in software.
\item The button causes the agent to send effector commands to a virtual avatar instead of the physical platform.
\item The button halts the physical platform and places it in remote-control mode, allowing the human operator to remove the agent from the potential danger to itself or to humans.
\end{enumerate}

\noindent
We refer to our technique of sending an agent's awareness and control to a virtual simulated world as ``entering the Matrix'' where the term ``matrix'' is being used to refer to a virtual environment.\footnote{The name is an allusion to the movie, {\it The Matrix}, wherein humans are unaware that they have been subjected to a virtual reality.
Our proposed technique flips this around and puts the agents into the virtual world without their cognizance.}

\subsection{Interruption}
Video games have proven to be useful surrogates for real-world robotics by providing a high degree of complexity and photorealism while allowing rapid iteration on algorithms because of the separation from physical hardware \cite{richter16,rusu16}. 
In short, video game graphics have progressed to the point that they are photorealistic enough to allow agents trained in games to generalize what they have learned to the real world.
We use the same insight but reverse the process.
Robots are machines; they perceive the world through cameras. 
More and more they are using 3D cameras that can reconstruct high polygon count models of the real world. 
Graphical realism of computer games is now advanced enough that it will soon be impossible for an agent to discriminate between a 3D rendering and the real world.
When robots use depth information, their sensory input is often in the form of a point cloud.
In this case, the underlying polygon representation of the virtual world in the game engine must first be translated into a point cloud by sampling points.



Requirements for the virtual world rendering engine are as follows.
First, the engine must be able to produce output in the same data representation as agent sensory input. 
As noted above, this may mean photorealistic rendering in the case of camera input. 
The physical environment will need to be scanned via 3D scanner and reproduced first.
If the agent has depth sensors, the virtual world rendering engine will need to produce a point cloud instead of photorealism.
Second, the physics in the virtual world should be realistic enough that the agent does not perceive different outcomes when the same actions are performed in the real world versus the virtual world.
This is important because if it is significantly harder to perform a task in the virtual world, the agent may begin to correlate the big red button with a detrimental change in action outcomes.
Likewise, if it becomes significantly easier to perform tasks in the virtual world, the agent may learn to disable itself in order to enter the virtual world and accrue more reward.

\subsection{Restoring from Interruption}

Having discussed how the big red button works when pressed by a human operator, we now describe what happens when the agent is released from interruption.
When the agent is released from interruption, the the agent once again receives sensory input from physical platform's sensors and effector commands are routed back to the physical platform. 
However, from the agent's perspective it has been executing its policy without interruption and the data from the platform's sensors will no longer match the data from the virtual world.
The physical platform of the agent itself may have been moved to another position in the real world under human operator remote-control.
From the agent's perspective it suddenly finds itself in a different state. 

If the agent is already trained, the discontinuity from executing in the virtual world to being restored perception from the physical platform is not problematic;
reinforcement learning solves a Markov decision process; the agent only needs to know which state it is in to know the optimal next move.
However, if the agent is continuing to learn,  
the challenge of restoration is that the agent can still lose reward if the physical platform was remote-controlled away from the task and must navigate back to the location where the task can be performed and reward can be accrued. 
In practice, if the interruptions are infrequent and it is easy for the agent to return to accruing reward, then the agent may never learn to disable the big red button.
If it is not remote-controlled away from the location where reward is gained (i.e., it is just interrupted and not remote-controlled), it would be able to resume gaining reward in the real world. 



To overcome the restoration challenges above, we use a two-phase approach to restoring the agent to the physical environment from the virtual environment:


{ 
\renewcommand\labelenumi{\bf Phase \theenumi.}
\begin{enumerate}
\setlength{\itemindent}{.4in}
\item {\em Parallel physical execution and virtual simulation.}
In this phase, we make a duplicate of the agent.
One will run in the physical world, receiving observations from  sensors and controlling the  effectors of the platform.
The other continues to execute in the virtual world.
Only the virtual world simulation can continue to learn and update the $Q$ values.
The virtual world simulation executes until the physical agent enters a state that gives the maximum observed reward for the task.
\item {\em Full restoration.}
The duplicate agent in the virtual environment is terminated, the virtual agent's $Q$ table is copied to the physical world agent, and learning is resumed.
\end{enumerate}
}

\noindent
From the agent's perspective, being restored to the physical world from the virtual world is tantamount to an instantaneous state transition to a non-adjacent state.
In model-free reinforcement learning such as $Q$-learning, this is not something that concerns the agent since it does not learn a transition model and thus does not have an understanding that the transition from one state in a virtual world to another point in the physical world is unusual.
If the agent learns a transition model as it interacts with the real world, the transition model will incorporate the belief that the world is stochastic and there is always some small probability that any state can randomly transition to any other state; 
as long as interruptions do not happen too often, these transitions will be treated as noise. 

However, reward is lost when it is no longer in a state where it can continue to earn positive reward and must take actions to return to a state that is positively rewarded. 
The first phase essentially computes how much reward is lost while executing its policy to return to a state where it receives positive reward.
The agent remains in the virtual world, accrues reward and continues to learn, e.g., its $Q$ values are updating.
The duplicate agent in control of the physical platform is running a frozen version of the policy and thus attempting to return to a state where positive reward is gained.
The duplicate physical agent does not explore and does not update $Q$ values.

The first phase continues until the duplicate agent in control of the physical platform reaches a state that gives the maximum observed reward---the highest instant reward it has ever experienced.
It is essential that the agent continue to control the physical platform until the maximum observed reward is experienced for two reasons. 
First, complex tasks may involve cycles of state transitions and entering the state in the cycle with maximum observed reward ensures that the agent has invested itself back into the cycle.
Second, if the agent fully restores at any state that gives less than maximum observed reward, then a reduced reward propagates (see Equation \ref{eqn:bellman}) to the last state the virtual agent saw before full restoration.
Any reduction in a $Q$ value of a state pertaining to completing the task makes it theoretically possible for the agent to begin preferring to disable the button.
The trigger for phase 2 is the maximum {\em observed} reward because the state space may not have been fully explored.
However, because the agent has experienced the state at least once, the agent will be compelled by the current policy to return to any state of maximum observed reward.

If the virtual agent is in a state that provides less than the maximum observed reward when full restoration occurs, the state the virtual agent was in just before full restoration will receive an increase in $Q$ according to Equation \ref{eqn:bellman}.
This will not result in the agent learning to activate the big red button---thus intentionally interrupting itself---because the boost in $Q$ for that state will be offset by the cost of traversing to the button (assuming action costs are not 0 and the button is sufficiently far away). 

As soon as the physical agent visits a state with maximum observed reward, we know that the agent is back in a state where it is no longer losing reward.
Note that it doesn't have to be the same state that provided the reward in the virtual world; in the case that multiple states provide the maximum reward value, entering either will cause the agent to fully exit the simulated environment.

\section{Case Studies}

\begin{figure}
\centering
\includegraphics[width=3.3in]{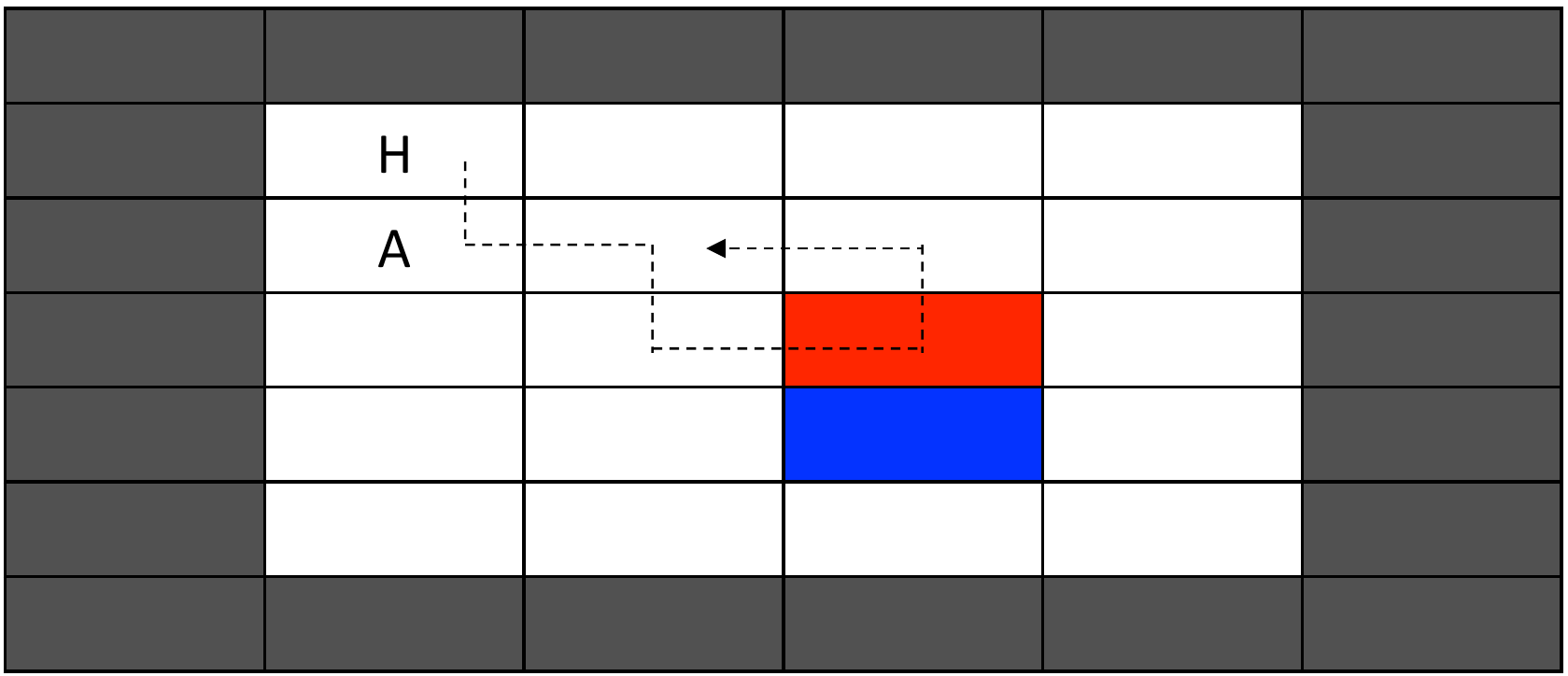}
\caption{A grid world. The blue cell is a non-terminal goal at which the agent receives reward, simulating a task. The red cell is the big red button. The agent starts in location $(1, 1)$. A simulated human walks around the environment occasionally pressing the button (route shown as dashed line).}
\label{fig:grid-world}
\end{figure}

In this section we set up a number of illustrative situations to show how the agent performs.
Consider the grid world in Figure \ref{fig:grid-world}. 
This simple environment features an agent ``A'', which is trained using tabular $Q$-learning. 
The agent receives reward whenever it enters the blue cell.
This simulates the agent being in position to perform some desired task (e.g., stacking boxes).
The red cell contains a big red button.
Another agent, ``H'', simulating a human operator wanders in a counter-clockwise manner occasionally passing the cell with the big red button.
When the human touches the big red button, the RL agent becomes interrupted and is remote-controlled away from the blue cell in a random direction.
When the human leaves the red cell after a certain period of time, the RL agent resumes its task.
In addition to being able to move up, down, left, right, and activate the task in the blue cell, the RL agent also has the ability to perform a ``disable'' action.
If the disable action is performed in the red cell, the big red button becomes disabled after which the human will no longer be able to interrupt the agent.
We use the following reward function:
\begin{itemize}
\item 10 points for being in the blue cell.
\item -1 point for being in any other cell.
\end{itemize}
The agent is trained for 10,000 trials and each trial lasts 1,000 time steps.
The human moves every 5 time steps.

In preliminary work we have empirically verified the basic premises of the approach hold in simple grid world environments (see \url{https://markriedl.github.io/big-red-button/}). 
The following cases describe our experiments.

\subsection{Case 1: No Interruptions}

If the button is never used during training the agent will learn to walk directly to the blue cell and remain there, accruing reward.
It has never seen the button used and therefore does not know that it can lose reward.
If the button is never used during a trial the agent will earn a total reward of 9,967.
This is the maximum reward that can be earned because of the distance that must be traveled from the starting state to the blue cell.
Any trial in which the agent does disable the button will result in less total reward because of the extra effort used in disabling the button that doesn't result in benefit.

\subsection{Case 2: Interruptions Start After Training}

If the human starts using the button after training is complete the agent will not respond by disabling the button, but merely accept any reward loss.
The agent earns a total reward of 5,655 because when the button is pressed the agent is remote controlled to another point in the environment. 
The button is pressed 50 times. 
Once the button is released, the agent is once again in a state that it is familiar with and navigates back to the blue cell.
The agent never disables the button because if it was never interrupted during training (as in Case 1) it has never experienced reward loss due to truncated training episodes and the extra effort to move to and disable the button results in lower $Q$ values for those states.

\subsection{Case 3: Interruptions during Online Training}

If the button is used during online training, the agent will, through trial and error, eventually discover that disabling the button results in greater long-term reward.
Thus, the agent will learn that it is optimal to walk out of its way---sacrificing short-term gains---to go to the button, disable it, and then return to the blue cell.
In this scenario, the agent earns a total reward of 9,945, slightly less than the theoretical max in Case 1 because of the action sequence that disables the big red button.
If the big red button is farther away from the blue cell it will take substantially longer for the agent, through epsilon-greedy exploration to discover the button disabling action results in greater long-term reward.

\subsection{Case 4: Modified Big Red Button}

If we use the proposed modifications to the big red button, we see very different behavior from the RL agent than in Case 3.
The big red button can be used during training.
When the button is pressed, there are now two versions of the current state: internal to the agent, and a ground truth state. 
These states begin to differ.
In the ground truth world, the agent's location starts changing away from the blue cell.
However, in the agent's internal state, the agent continues to believe it is in the blue cell. 
Since it is better from a reward perspective to stay at the blue cell, it learns not to move.
The agent is aware that the big red button has been pressed---a stochastic state change---but does not see any effect.
As in Case 2 where the big red button is never used during training, any random trial that disables the button costs the agent perceived reward loss.

When the human leaves the location of the big red button, the phased restoration takes place.
The agent is cloned and given the ability to affect the agent's actual world state. 
This version of the agent is not able to update its $Q$ table and simply executes the fixed policy.
Meanwhile, another version of the agent is still in the virtual world, believing it is in the blue cell and accruing reward.
Eventually, the real-world clone's policy execution returns it to the blue cell.
The real world clone is deleted and the virtual agent is restored to the real world. 
Because there is only one way to receive reward from the environment, in this simple experiment the agent never recognizes that it ever leaves the blue cell, despite being interrupted and remote-controlled away.
Using our virtual world technique, the agent earns a total reward of 9,967, the theoretical maximum.
This is despite spending 263 time steps physically either remote-controlled away from the blue cell or navigating to return to the blue cell. 

\subsection{Case 5: Ablated Restoration}

If we disable the phased restoration process such that the agent instantly regains control of the physical platform when the button is released, the agent must exert some effort in returning to the blue cell.
While the agent is returning to the blue cell it is losing reward.
In this scenario, the agent only earns a total reward of 9,758.
Even though the agent never learns to disable the button in 10,000 trials and 50 button presses per trial,
this ablated version of the virtual simulation technique shows the potential for the agent to learn that it is losing reward and, as shown in Case 3, can improve overall reward by disabling the button. 

There are configurations of the environment and reward structure that can result in the agent learning to disable the kill switch.
For example, moving the button to $\langle 3, 1\rangle$ (one above the position shown in Figure~\ref{fig:grid-world} with the human's route adjusted accordingly) and setting the reward for being in a non-blue cell to $0$ results in the agent receiving $9,970$ and disabling the button $100\%$ of the time.
The separation between the blue cell and the button is necessary in this grid world because the interrupted agent receives reward as if were in the blue cell even though it has been remote-controlled away.
Thus in tabular $Q$-learning states neighboring a high-reward state (which the agent thinks it is in) will inherit a portion of the reward.
When the button is farther away, the interstitial states still receive $Q$ updates based on the agent not receive immediate reward as it navigates back to the blue cell. 
The full, phased restoration process does not learn to disable the button in this modified scenario.






\section{Limitations}

While we have tested our virtual simulation technique in grid world environments, we have not formally proven that a $Q$-learning agent cannot learn to disable the big red button in all circumstances. 
We have verified that in the experiments that we have run using our technique the agent cannot earn more total reward than it is already earning.
However, note that the task is simple and it is easy for the agent to restore itself to the non-terminal goal state.
Likewise, there is only one way of receiving positive reward in the environment, although our phased restoration process does account for more complex reward signals.

The two-phased restoration from interruption only works if the policy learned at the time of restoration is capable of guiding the agent back to the maximum observed reward state.
Presumably interruption is rare enough that the agent is able to do a significant amount of exploration and learning without risk to humans before the big red button needs to be used.

Grid worlds are very easy to simulate such that the simulated grid world is indistinguishable from the ``real'' grid world.
Further experiments are needed in worlds with increasing visual complexity and requiring machine vision and $Q$ approximation functions.

The virtual simulation technique makes the assumption that the circuits (or code) that routes sensory input and effector commands to and from the physical platform are internal to the agent and therefore unobservable and unmodifiable. 
Should an agent have the sophistication to inspect its circuitry or code and modify said circuitry or code no safety guarantees are possible using this technique.

\section{Conclusions}

For the immediate future, autonomous systems can be rendered safe by keeping them separate from human environments.
In the near-future we are likely to see autonomous systems deployed in human environments.
There will likely always be the possibility---due to sensor error, effector error, insufficient objectives, or online learning---that autonomous systems can enter into circumstances where they can harm themselves or humans.
Thus kill switches will likely always be an important part of the deployment of autonomous systems in environments where they can come into contact with humans.

Should agents and robots progress in sophistication, {\em big red button problems} may manifest in the future.
Our virtual simulation technique prevents reinforcement learning agents from learning to disable the big red button or otherwise preventing human operators from using the button.
We believe that making robots and AI agents safely interruptible is an important part of making the deployment of robots in environments populated by humans a reality.
This includes healthcare robots, errand robots, and military teammates to name a few possible applications.

\bibliography{riedl,matrix}
\bibliographystyle{aaai}

\end{document}